\documentclass{article}

\usepackage{PRIMEarxiv}

\usepackage{soul}
\usepackage{url}
\usepackage{hyperref}
\usepackage[utf8]{inputenc}
\usepackage{caption}
\usepackage{xcolor}
\usepackage{colortbl}
\usepackage{graphicx}
\usepackage{caption}
\usepackage{courier}
\usepackage{multirow}
\usepackage{subcaption}
\usepackage{color}
\usepackage{todonotes}
\usepackage{amssymb}
\usepackage{gensymb}
\usepackage{enumitem}
\usepackage{booktabs}
\usepackage[linesnumbered, ruled]{algorithm2e}
\usepackage{amsmath,amssymb}
\usepackage[subtle]{savetrees}
\usepackage{algorithmic}
\usepackage{pifont}
\usepackage{xcolor}
\usepackage{etoolbox}

\usepackage{soul}
\usepackage{url}
\usepackage{hyperref}
\usepackage[utf8]{inputenc}
\usepackage{caption}
\usepackage{xcolor}
\usepackage{colortbl}
\usepackage{graphicx}
\usepackage{caption}
\usepackage{courier}
\usepackage{multirow}
\usepackage{subcaption}
\usepackage{color}
\usepackage{todonotes}
\usepackage{amssymb}
\usepackage{gensymb}
\usepackage{enumitem}
\usepackage{booktabs}
\usepackage[linesnumbered, ruled]{algorithm2e}
\usepackage{amsmath}
\usepackage{xcolor}

\newcommand{\ours}{{\textsf{EaP}}\xspace}
\newcommand{\production}{{\textsf{Prod}}\xspace}

\newcommand{\tone}{\textsf{Navigation}\xspace}
\newcommand{\ttwo}{\textsf{Order}\xspace}
\newcommand{\tthree}{\textsf{Deal}\xspace}
\newcommand{\tfour}{\textsf{Q2Q}\xspace}

\title{Examples as the Prompt: A Scalable Approach for Efficient LLM Adaptation in E-Commerce}

\author{
  Jingying Zeng$^*$, Zhenwei Dai$^*$, Hui Liu,  Samarth Varshney,  \\
  \textbf{Zhiji Liu},  \textbf{Chen Luo}, \textbf{Zhen Li}, \textbf{Qi He}, \textbf{Xianfeng Tang}$^*$ \\
  Amazon \\
  \texttt{xianft@amazon.com} \\
}

\begin{document}
\maketitle
\def\thefootnote{}\footnotetext{$^*$ Equal contribution.}

\begin{abstract}
    Prompting LLMs offers an efficient way to guide output generation without explicit model training. In the e-commerce domain, prompting-based applications are widely used for tasks such as query understanding, recommender systems, and customer support. However, adapting LLMs to different tasks often requires extensive prompt engineering by domain experts, along with frequent updates to align with evolving business needs. Additionally, crafting fully unbiased natural language prompts remains a challenge for humans.
    To address these challenges, we propose a novel framework, \underline{E}xamples \underline{a}s the \underline{P}rompt (\ours) 
    which leverages labeled data to enhance prompts. Specifically, \ours automatically selects the most representative examples to maximize the few-shot capability of LLMs. It is efficient due to its unsupervised example selection and adaptive to potential data distribution shifts.
    We validate \ours on four real-world production use cases, demonstrating that it achieves comparable or even superior performance comparing to hand-crafted prompts designed by domain experts. Additionally, we introduce $\ours_{lite}$, which entirely replaces the natural language components of prompts with labeled examples. $\ours_{lite}$ improves LLM inference speed by up to 70\% without compromising performance.
    Latest online A/B test shows that using \ours and $\ours_{lite}$ for data labeling can bring significant composite revenue gain  by 0.06\%.
    
\end{abstract}


\section{Introduction}

With the increasing popularity of large language models (LLMs), more and more prompt-based applications are deployed into e-commerce productions such as query understanding \cite{anand2023query}, recommender system \cite{liu2024sequential}, and customer support \cite{su2025llm}. Comparing to finetuning that requires substantial computational resources and time to update LLM's parameters, prompting LLMs enables rapid iteration and adaptation to new customer experiences by simply adjusting the input prompts \cite{shin2023prompt,brown2020language}. This flexibility is particularly advantageous in testing new ideas and minimum viable products, where requirements frequently change. Generally speaking, the main part of a prompt is natural language task definition. Some prompts also include a few in-context examples to illustrate the task \cite{dong2022survey,li2024long}.

However, designing, improving, and maintaining prompt-based applications still suffer from several limitations: 
(1) Prompts need regular updates to reflect the latest business needs. Most prompts require extensive hand-crafting by domain experts. To ensure the freshness of the prompts, domain experts need to update the outdated prompt periodically, which is time-consuming and labor-intensive \cite{kong2024prewrite};
(2) It is very challenging to adopt human annotated data into the prompt. Due to runtime constraints, the length of prompt is limited \cite{liu2025effects}. Therefore, adding more and more examples into the prompt to cover corner cases is not practical. 
(3) it is extremely difficult for human to define/describe the business needs with purely unbiased natural languages \cite{goldfarb2023prompt,xu2024take}. For example, when both “very important” and all-caps are used to emphasize constraints, it can be difficult to determine which conveys greater importance. Moreover, it is challenging to consistently select in-context examples that are both unbiased and sufficiently representative for the application.

Motivated by recent advances in in-context learning (ICL) \cite{brown2020language,wei2022emergent,zhou2023mystery}, we aim to address the above challenges by leveraging both LLMs' emerging ability to learn in context, and the rich real-traffic data collected from e-commerce applications. Several studies analyze the importance of in-context example selections to the LLMs' ICL performance \cite{wang2023large,li2024implicit}. In light of this, instead of improving the natural language instructions in the prompt, ICL leverages labeled data as examples, such that the few-shot capability of LLMs are maximized to perform specified tasks~\cite{wang2023learning}. 
We propose a novel framework \underline{E}xamples \underline{a}s the \underline{P}rompt (\ours) that automatically retrieves the optimal in-context examples that cover both corner cases and data distribution shift ~\cite{wei2023larger}.
\ours constructs the actual prompt on-the-fly with both natural language instructions and dynamic in-context examples selected based on the actual input to the application (e.g., search query). In particular, both global examples representing the task and local examples being similar to the specific input are selected as in-context examples. 
\ours is unsupervised with no training needed. \ours is flexible and can adopt different retrieval algorithms and similarity measurements. We further introduce $\ours_{lite}$ which completely replaces the system prompts with labeled examples. \ours and $\ours_{lite}$ simplify prompt management and allows us to focus on data annotation quality. 

We validate \ours on four e-commerce tasks. \ours consistently shows comparable and even better performance comparing to the original hand-crafted prompt in production. $\ours_{lite}$ significantly reduces the runtime by up to 70\%. 
We test \ours and $\ours_{lite}$ in data labeling pipeline where online A/B test suggests the potential of improvement on composite revenues by 0.06\%.


\section{Methodology}
In e-commerce, LLMs are prompted to generate certain output. For example, in a query parsing task, LLMs take the search query of a user as input, and output the parsing results (e.g., keywords, brand, size, etc.) as json formatted text string. Let $q$ denote the input and $o$ denote the output for the LLMs, we first fit $q$ into a natural language prompt template (i.e., system prompts), then let LLMs read the prompt formatted with $q$ to produce the desired output $o$. 

\begin{figure}[t]
    \centering
    \includegraphics[width=0.65\linewidth]{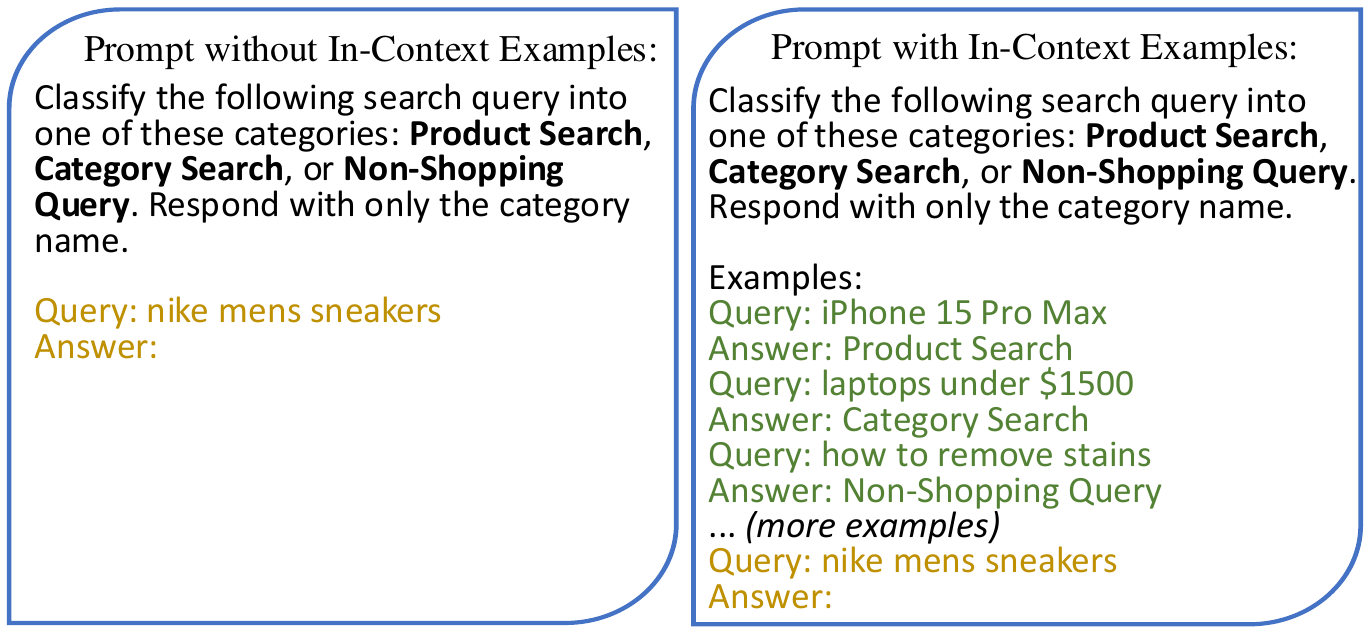}
    \caption{Prompt examples for search query classification. The green text represents in-context examples, while the yellow text denotes the input $q$.}
    \label{fig:prompt_examples}
\end{figure}

Because many e-commerce tasks are less common in open-source domain, prompt engineers would hand-pick some representative examples for the specific task, and compile them into the system prompt. These in-context examples stimulate the few-shot/zero-shot capabilities of LLMs and improve performances. In most cases, these examples are selected from a large labeled data source accumulated from real business. For example, queries and their human audited/annotated parsing results can be treated as a data source. Figure \ref{fig:prompt_examples} illustrates the same system prompt without and with in context examples. 

Let $\mathcal{E}$ denote the whole labeled data source, \ours aims to find a small set of most representative examples $\mathbf{E} = \{e_1, e_2, \cdots, e_n\} \in \mathcal{E}$ to maximize the few-shot capability of LLMs. In particular, \ours selects examples from both global and local perspectives.

\begin{table*}[htb]
\centering
\small
\caption{Comparison of \ours and production prompts \production.}\label{tab:overall_comp}
\begin{tabular}{c|cccc|cccc|cc|cc}
\toprule
\multirow{2}{*}{} & \multicolumn{4}{c|}{\tone}                                           & \multicolumn{4}{c|}{\ttwo}                                          & \multicolumn{2}{c|}{\multirow{2}{*}{\tthree}} & \multicolumn{2}{c}{\multirow{2}{*}{\tfour}} \\
                           & \multicolumn{2}{c}{Positive}        & \multicolumn{2}{c|}{Negative}       & \multicolumn{2}{c}{Positive}        & \multicolumn{2}{c|}{Negative} & \multicolumn{2}{c|}{}                      & \multicolumn{2}{c}{}                     \\ 
                    & Prec.        & Recall           & Prec.        & Recall           & Prec.        & Recall           & Prec.          & Recall   & Prec.               & Recall           & BLEU                & ROUGE              \\ \midrule
 \production & 0.8198          & 0.9040          & 0.9565          & 0.9140          & 0.9916           & 0.9516          & 0.9720            & 0.9952  & 0.9500                    & 0.9717          & 0.7736             & 0.8088            \\
 \ours       & \textbf{0.8910} & \textbf{0.9205} & \textbf{0.9651} & \textbf{0.9513} & \textbf{0.9918} & \textbf{0.9731} & \textbf{0.9843}   & 0.9952  & \textbf{0.9715}        & \textbf{1}       & \textbf{0.7957}    & \textbf{0.8136} \\
\bottomrule
\end{tabular}
\end{table*}

\subsection{Global Examples}\label{Global}

Global examples represents the data distribution of the entire problem space for the task. By including global examples in the prompt, we help LLMs understand the problem and learn the correlations between input and output. Ideally, the more representative global examples are chosen, the better performance the prompt can achieve.
Global examples can be retrieved through unsupervised methods, such as random selection and K-means clustering.
However, when $\mathcal{E}$ is not a uniform sample of the problem space, or when $\mathcal{E}$ contains imbalanced distributed classes, subsampling and/or downsampling techniques can be adopted to reduces the biases in global example selection. 
When the task is classification, balancing the number of examples from every class can also mitigate the imbalance issues.
Because global examples are the same for every input, we can further optimize their representativeness with data condensation approaches \cite{wang2023large, li2024implicit}. Specifically, a small set $\mathbf{E}^*$ where $ |\mathbf{E}^*| = n$ is learned/optimized on $\mathcal{E}$ such that any machine learning model trained on $\mathbf{E}^*$ can achieve the same or very similar performance comparing to the same model trained on $\mathcal{E}$.
In practice, replacing manually selected example in the prompt with global examples selected by \ours almost always improves performance. In addition, \ours can update global examples frequently, especially for newly enabled features where user behaviors evolves relatively fast.


\subsection{Local Examples}\label{Local}
While global examples can already significantly improves the performance, we noticed prompt-based applications could still make mistakes, especially when the input $q$ is less representative.
To prevent LLMs from overlooking it, we find examples from $\mathcal{E}$ that are most similar to the specific input, and include them as local examples in $\mathbf{E}$.
We first featurize $\mathcal{E}$ with bag of words, TF-IDF, and/or word embeddings, then apply corresponding similarity measurements such as Euclidean distance, Jaccard similarity, and/or cosine similarity. Because local examples are selected in real-time, we further adopt approximate nearest neighbor search methods (e.g., Faiss) to reduce its runtime.

\begin{figure}[t]
    \centering
    \includegraphics[width=0.45\linewidth]{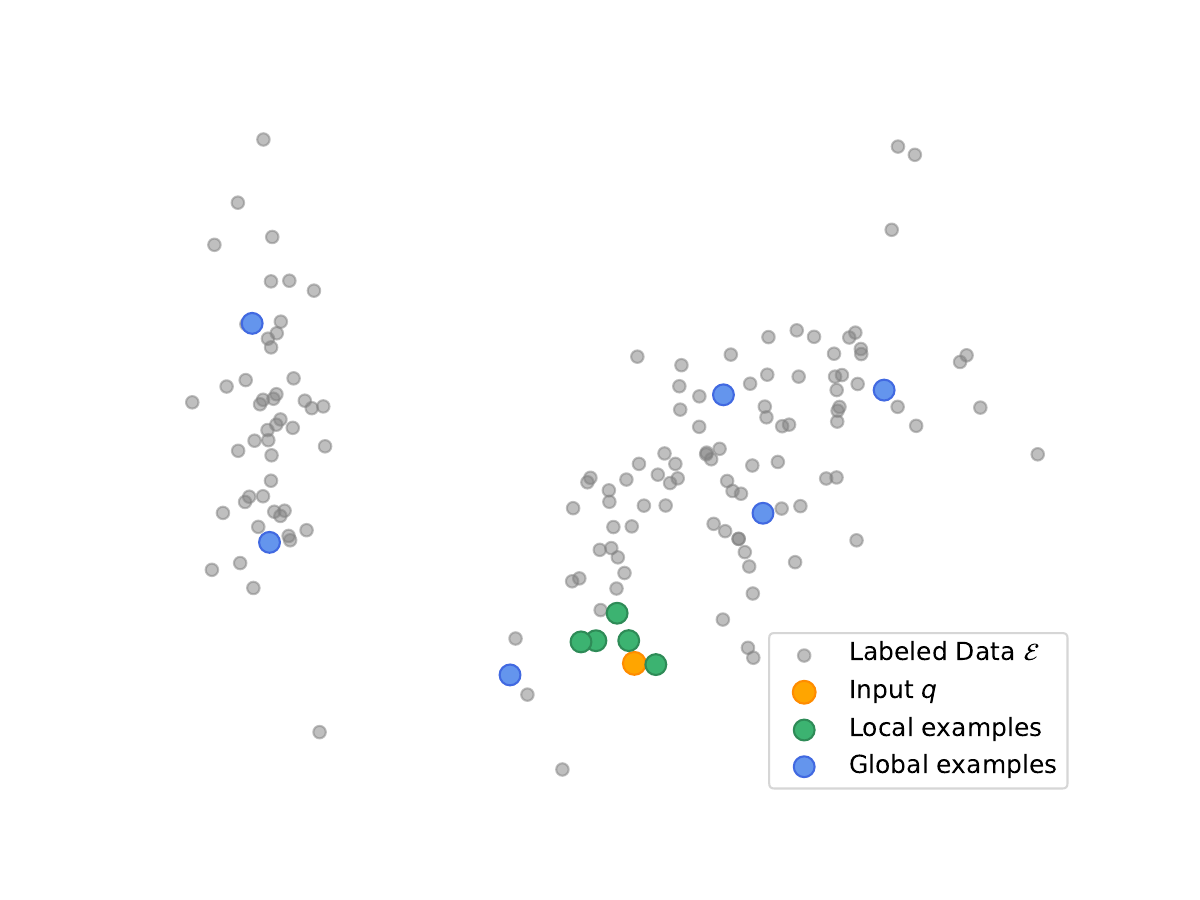}
    \caption{Visualization of global and local examples selected by \ours. While global examples represents the overall data distribution, local examples help characterize the specific input.}
    \vspace{-2em}
    \label{fig:global_local_examples}
\end{figure}

While global examples provide an overview of the problem space, local examples sharp the boundaries between sub-problems and prevent LLMs from being over-confident. Local examples are particularly helpful when $q$ is a corner case or near the boundary of two classes. We illustrate the difference between global and local examples in Figure \ref{fig:global_local_examples}.

\subsection{Performance v.s. Runtime}
Comparing to examples in Figure \ref{fig:prompt_examples}, system prompts in real production system can be relatively longer with task definition, formatting instruction, and safety guardrails. Since longer prompts lead to longer inference time for LLMs, the latency constraints limit the total length and total number of in-context examples when deploying prompt-based applications.
In practcie, we notice that part of the system prompt can be removed when sufficient in-context examples are added by \ours. In most cases, replacing natural language task definition with examples could shorten the prompt and reduce the inference latency for LLMs hosted online.

When latency is not a constraints, adding more examples without modifying existing prompt could achieve the best performance. For offline tasks such as data labeling and generation, we can tune the number of examples in \ours to maximize the performance of prompt-based applications.

\section{Experiments}
We aim to answer the following research questions through our experiments:
\begin{itemize}[leftmargin=*]
    \item \textbf{RQ 1}: Can \ours out-perform state-of-the-art production prompt engineered by human expert?
    \item \textbf{RQ 2}: How does global and local examples impact the performance of \ours?
    \item \textbf{RQ 3}: How \ours reduces the inference runtime of LLMs by shorten the prompts?
\end{itemize}

\subsection{Setup}
Four real e-commerce tasks with prompt-based solutions are selected to validate \ours:
\begin{enumerate}[leftmargin=*]
    \item \textbf{\tone}: In addition to shopping for products, more and more users tend to search for the functionality they need directly in search box. For example, users would search ``customer support'' directly, rather than navigate to the customer support page. We aim to classify the intent of search queries between navigation and shopping. We collect 730 navigation queries and 1554 regular shopping queries as the labeled.
    \item \textbf{\ttwo}: We further want to identify if a search query is clearly asking about a past order placed by the user. We formulate it as a binary classification problem. We collect 1000 past order related queries and 1000 non-order related queries as the labeled dataset.
    \item \textbf{\tthree}: Detecting deal related search intent is critical for e-commerce websites. We collect 2457 search queries with 492 being labeled as deal intent.
    \item \textbf{Question2Query} (\tfour): \tfour transforms question-style search queries into non-question, shoppable formats. For example, it converts ``how to clean a cast iron skillet'' into ``cast iron skillet cleaning.'' \tfour enhances the coverage and effectiveness of traditional e-commerce search functions, such as rule-based solutions, before transitioning to modern machine learning and LLM-based approaches. We collect 1543 question-style queries paired with their corresponding shoppable counterparts.
\end{enumerate}

We randomly split the collected labeled data equally into validation set and test set. We use validation set as the labeled data $\mathcal{E}$, and compare \ours against our state-of-the-art human engineered prompts in production. Precision and Recall metrics are reported on \tone, \ttwo and \tthree; while BLEU \cite{papineni2002bleu} and ROUGE \cite{lin2004rouge} are reported for \tfour.

Without the loss of generality, we adopt Mistral-7B-Instruct-v0.2\footnote{https://huggingface.co/mistralai/Mistral-7B-Instruct-v0.2} \cite{jiang2023mistral} with vLLM \cite{kwon2023efficient} for our experiment. To ensure a fair comparison, the total number of examples selected by \ours are always equal to the number of examples in production prompts on every task. We use sklearn $K$-means clustering\footnote{https://scikit-learn.org/stable/modules/generated/sklearn.cluster.KMeans.html} to select global examples, and use RapidFuzz\footnote{https://github.com/rapidfuzz/RapidFuzz} to find local examples in our offline evaluation.
All experiments are conducted on a single AWS EC2 P4 Instances with A100 GPUs. Note that GPU is only used for LLM inference and example selection in \ours are conducted on CPU only.


\subsection{Performance Analysis}

We report the performance comparison of \ours and \production in Table \ref{tab:overall_comp}. Based on the results, we observe that \ours can consistently outperform the handcrafted production prompts over all the tasks, which shows the advantages of smartly selecting few-shot examples in ICL over static human-written prompt. \ours achieves significant improvements on detecting navigational intent in the task \tone. One potential reason is that \tone is hard to be defined precisely via natural language definition. Leveraging representative global and local examples from \ours could mitigate the gap in system prompt and improve the performance. For other tasks such as \ttwo, \tthree, and \tfour, \ours  achieves slightly better performance than the \production prompt. However, \ours shortens the prompt lengths by eliminating the length natural language instructions from system prompt, which can significantly reduce the latency during online service. In Section \ref{sec:global_and_local}, we will have a more thorough analysis on the inference speed comparison.


\subsection{Global and Local Examples}\label{sec:global_and_local}
For a fixed budget of total number of examples, we want to study how to balance the number of global examples and local examples to maximize the performance. Taking \tone as an example, we keep the total number of in-context examples unchanged, while alter the number of global and local examples. The prompt used in production includes $11$ in-context examples. To have a fair performance comparison to the production setup, we also set the total number of in-context examples to $11$.
The precision and recall on both positive and negative class for navigational intent classification are reported in Table \ref{tab:global_local_tuning}.
We notice the both precision and recall will first increase and then decrease, showing that balanced numbers of global and local examples are important for \ours.
Moreover, we observe that the positive class is more sensitive to the change, due to the fact that it is the minority class in this unbalanced dataset. Selecting the right balance of global and local examples is crucial to \ours's performance.

\begin{table}[h!]
\centering
\caption{Changing the number of global and local examples.}\label{tab:global_local_tuning}
\small
\renewcommand\arraystretch{1}
\begin{tabular}{cc|cccc}
\toprule
\# global  & \# local & \multicolumn{2}{c}{Positive}        & \multicolumn{2}{c}{Negative}        \\
                  examples                  &            examples                        & Precision        & Recall           & Precision        & Recall           \\ \midrule
11                                  & 0                                  & 0.8653          & 0.8264          & 0.9232          & 0.9419          \\
10                                  & 1                                  & 0.8134          & 0.8971          & 0.9213          & 0.9071          \\
9                                   & 2                                  & 0.8066          & 0.8714          & 0.9398          & 0.9057          \\
8                                   & 3                                  & 0.8613          & 0.8585          & 0.9362          & 0.9376          \\
7                                   & 4                                  & 0.8696          & 0.8360          & 0.9273          & 0.9434           \\
6                                   & 5                                  & 0.8850          & 0.8907          & 0.9505          & 0.9478          \\
5                                   & 6                                  & \textbf{0.9010} & \textbf{0.9205} & \textbf{0.9651} & 0.9513          \\
4                                   & 7                                  & 0.9066          & 0.8424          & 0.9311          & \textbf{0.9608} \\
3                                   & 8                                  & 0.8822          & 0.8907          & 0.9504          & 0.9463           \\
2                                   & 9                                  & 0.8300          & 0.8778          & 0.9648          & 0.9144          \\
1                                   & 10                                 & 0.8112          & 0.8842          & 0.9455          & 0.9071          \\
0                                   & 11                                 & 0.7881          & 0.8971          & 0.9505          & 0.8912   \\
\bottomrule
\end{tabular}
\end{table}



\subsection{Effectiveness of System Prompt}
For the classification tasks like \tone, LLM heavily relies on the in-context examples to make the prediction, while the system prompt can only provide general and high-level guidance. In this experiment, we investigate to which extend the role of system prompt can be replaced by the in-context examples.
Surprisingly, we found that even when removing the system prompt and replacing it with a simple one-line instruction:
\begin{center}
    \textbf{\textit{Here are some examples of expected input and output:}}
\end{center}
The LLM still achieves similar performance, provided sufficient in-context examples. We create an ablation $\ours_{lite}$ with such one-line system prompt and study how its performance would change w.r.t the number of examples on Task \tone.
We fix the ratio of global and local examples according to the learning in \ref{sec:global_and_local}. 
We compare the inference speed of each method based on the following two two metrics: (1) \textbf{Item per second (it/s)}: this metrics quantifies the number of prompts that can be processed per second. (2) \textbf{Output token per second (toks/s)}: this metrics quantifies how many output tokens are generated per second. 
We report the overall accuracy (i.e., if a query intent is correctly classified) in Table \ref{tab:eap_lite}. The performance of \ours converges after the total number of examples reaches about 11. Even though the inference speed decreases almost linearly with the addition of more examples, \ours still runs faster than the hand-crafted prompt \production and \ours, even when $39$ examples are used. This motivate us to supress and/or remove system prompt when possible in production.

\begin{table}[h!]
\centering
\small
\caption{$\ours_{lite}$ outperforms \production with no system prompt.}\label{tab:eap_lite}
\renewcommand\arraystretch{0.9}
\begin{tabular}{ccc|ccc}
\toprule
\# global   & \# local  & Total \#  & Accuracy & Input & Output  \\
examples  & examples  & samples &        &  it/s & toks/s \\
\midrule
1                   & 2                  & 3                & 0.819    & 48.11        & 3,079.31      \\
2                   & 3                  & 5                & 0.884    & 45.71        & 2,925.23      \\
3                   & 4                  & 7                & 0.911    & 44.07        & 2,820.79      \\
4                   & 5                  & 9                & 0.928    & 41.90        & 2,681.76      \\
5                   & 6                  & 11               & 0.932    & 40.11        & 2,566.81      \\
6                   & 7                  & 13               & 0.933    & 38.93        & 2,491.43      \\
7                   & 8                  & 15               & 0.938    & 37.40        & 2,393.90      \\
8                   & 9                  & 17               & 0.948    & 35.72        & 2,286.09      \\
9                   & 10                 & 19               & 0.941    & 34.28        & 2,194.15      \\
10                  & 11                 & 21               & 0.947    & 32.07        & 2,052.43      \\
11                  & 12                 & 23               & 0.942    & 31.90        & 2,041.87      \\
12                  & 13                 & 25               & 0.942    & 30.65        & 1,961.50      \\
13                  & 14                 & 27               & 0.957    & 29.87        & 1,911.49      \\
14                  & 15                 & 29               & 0.957    & 28.57        & 1,828.71      \\
15                  & 16                 & 31               & 0.942    & 27.80        & 1,779.30      \\
16                  & 17                 & 33               & 0.943    & 26.84        & 1,717.46      \\
17                  & 18                 & 35               & 0.948    & 26.37        & 1,687.84      \\
18                  & 19                 & 37               & 0.950    & 25.58        & 1,636.85      \\
19                  & 20                 & 39               & 0.944    & 25.25        & 1,616.19      \\
\midrule
\multicolumn{2}{c}{\production} & 11               & 0.910    & 23.16        & 1,482.20   \\
\multicolumn{2}{c}{\ours (5 global + 6 local)} & 11               & 0.932 & 23.16        & 1,482.20   \\
\bottomrule
\end{tabular}
\end{table}

\section{Production Usecases}
\ours can be widely adopted to improve data labeling and assist human auditing. With a small set of high-quality human annotated data, we can leverage \ours to quickly expand them to a much larger LLM-labeled dataset. We report the efficiency improvement by adopting $\ours_{lite}$ into the data labeling pipelines for task \tone, \ttwo, and \tthree in Table \ref{tab:data_labeling_speed} using the same offline datasets.
Clearly, $\ours_{lite}$ significantly increases the output token per second speed by up to 70\%, due to the shorter prompts.

\begin{table}[h!]
\centering
\caption{$\ours_{lite}$ imporves data labeling efficiency.}\label{tab:data_labeling_speed}
\small
\begin{tabular}{c|ccccccc}
\toprule
\multirow{2}{*}{}                                                          & \multicolumn{2}{c}{\tone} & \multicolumn{2}{c}{\ttwo} & \multicolumn{2}{c}{\tthree}  \\
                                                                           & it/s           & toks/s         & it/s            & toks/s         & it/s        & toks/s              \\ \midrule
\production                                                                & 23.16          & 1482.20         & 27.19           & 1740.45        & 25.36       & 1623.32             \\
$\ours_{lite}$                                                                        & 39.74          & 2543.59        & 44.42           & 2842.74        & 41.3        & 2643.44             \\
\midrule
\begin{tabular}[c]{@{}c@{}} Improvement\end{tabular} & 71.59\%        & 71.61\%        & 63.37\%         & 63.33\%        & 62.85\%     & 62.84\%             \\
\bottomrule
\end{tabular}
\end{table}

Downstream applications can utilize extra data labeled by \ours for their model training. One example is expanding labeled dataset size for underrepresented classes. For example, recognizing new brands is relatively more challenging  than their well-established counterpart. We leverage \ours to expanding the labeled data for Brand Named entity recognition (NER) model. A new model trained on expanded labeled data have been tested via online A/B test. In particular, the A/B test results shows the potential of significant gains on the top-line metrics composite revenues  by 0.06\%.

\section{Conclusion}
This paper introduces a novel approach, \ours, to select representative in-context learning examples that can optimize the LLM prediction/generation quality. Compared to static, human-crafted prompts, \ours eliminates the need for costly human effort while achieving superior performance. Further analysis reveals that the example-only prompt, which replaces the human-designed system prompt with a simple one-line instruction, maintains comparable performance while significantly reducing latency. We envision that our approach can be implemented across various domains, fostering faster development of LLM applications in industry.


\bibliographystyle{unsrt}  
\bibliography{references}

\end{document}